\pdfoutput=1
\documentclass[11pt]{article}

\usepackage[T1]{fontenc}
\usepackage[utf8]{inputenc}
\usepackage{lmodern}
\usepackage[margin=1in]{geometry}

\usepackage{natbib}
\usepackage[hidelinks]{hyperref}

\usepackage{microtype}
\usepackage{graphicx}
\usepackage{subfigure}
\usepackage{booktabs} 
\usepackage{comment}
\usepackage{algorithm2e,setspace}
\usepackage{etoc}
\usepackage{titletoc}
\usepackage{shorttoc}
\usepackage{titlesec}
\usepackage[toc,page,header]{appendix}
\usepackage{minitoc}
\usepackage{fancyvrb}
\usepackage{lineno}
\usepackage{amsmath}
\usepackage{amssymb}
\usepackage{mathtools}
\usepackage{amsthm}
\usepackage{enumitem}
\usepackage{wrapfig,lipsum,booktabs}
\usepackage{caption}
\usepackage{adjustbox}
\usepackage{textcomp}
\usepackage{pythonhighlight}
\usepackage{algorithmic}
\usepackage{xcolor}

\theoremstyle{plain}
\newtheorem{theorem}{Theorem}[section]

\theoremstyle{definition}
\newtheorem{definition}[theorem]{Definition}

\theoremstyle{remark}

\DeclareMathOperator*{\argmax}{arg\,max}

\def\*#1{\mathbf{#1}}
\def\+#1{\mathcal{#1}}
\def\emp*#1{{#1}_n}

\definecolor{blush}{rgb}{0.87, 0.36, 0.51}
\definecolor{ForestGreen}{RGB}{34, 139, 34}

\newcommand{\ERED}[1]{\textcolor{red}{\boldsymbol{#1}}}
\newcommand{\EBGE}[1]{\textcolor{blue}{\boldsymbol{#1}}}
\newcommand{\EGGE}[1]{\boldsymbol{\textcolor{ForestGreen}{#1}}}

\title{LATTS: Locally Adaptive Test-Time Scaling}

\author{
    Theo Uscidda$^{1,}$\thanks{Work done during an internship at AWS AI Labs.} \,\,\,
    Matthew Trager$^{2,}$\thanks{Correspondence to \texttt{mttrager@amazon.com}} \,\,\,
    Michael Kleinman$^{2}$ \,\,\,
    Aditya Chattopadhyay$^{2}$\\[.8em]
    Wei Xia$^{2}$\qquad \qquad 
    Stefano Soatto$^{2}$\\[1.5em]
    $^{1}$CREST-ENSAE, IP Paris \qquad $^{2}$AWS AI Labs\\ 
}
\date{}

\begin{document}
\maketitle

\begin{abstract}
One common strategy for improving the performance of Large Language Models (LLMs) on downstream tasks involves using a \emph{verifier model} to either select the best answer from a pool of candidates or to steer the auto-regressive generation process towards better outputs. This class of methods typically results in improved accuracy at the cost of increased computation at test-time, a paradigm known as \emph{test-time scaling}. However, most existing approaches increase computation uniformly across all samples and generation steps, without considering the complexity of individual instances, leading to inefficient resource use. We address this limitation by proposing an approach, called \emph{Locally Adaptive Test-Time Scaling (LATTS)}, that allocates variable compute across generation steps. Specifically, at each generation step, LATTS employs a verifier-based acceptance criterion to decide whether to resample, backtrack, restart, or stop the generation process. This criterion effectively adjusts the per-step computational effort based on a precise notion of \emph{local difficulty} derived from the verifier model. Empirical results show that LATTS achieves significantly superior accuracy--compute tradeoffs compared to standard verifier-based methods. 
\end{abstract}

\section{Introduction}

Scaling computation at test time has emerged as a powerful complementary strategy to traditional scaling of training resources~\citep{snell2024scalingllmtesttimecompute,liu2025bagtricksinferencetimecomputation,liu20251bllmsurpass405b,muennighoff2025s1simpletesttimescaling}. By allocating additional computation during inference, models can engage in deeper reasoning and tackle increasingly complex tasks. Recent examples such as DeepSeek R1~\citep{guo2025deepseek} or Kimi K1.5~\citep{kimiteam2025kimik15scalingreinforcement} 
demonstrate substantial performance improvements on challenging problems such as mathematical reasoning when given increased inference budgets. Typically, these methods rely on prompting the model to reason step-by-step, a technique known as \emph{Chain-of-Thought (CoT)} prompting~\citep{wei2023chainofthoughtpromptingelicitsreasoning,nye2021workscratchpadsintermediatecomputation,zhou2023leasttomostpromptingenablescomplex}, wherein each intermediate generated ``thought'' corresponds to a discrete reasoning step.

In addition to sequentially scaling computation, a growing body of research focuses on guiding the reasoning process using external verifiers,particularly \emph{Process Reward Models (PRMs)}\citep{lightman2023letsverifystepstep}. PRMs assess the quality of a reasoning chain by generating a sequence of scores that indicate the likelihood of correctness at each intermediate step. This enables the design of fine-grained search algorithms, like Monte-Carlo Tree Search or Beam Search, which use the feedback from step-level PRM scores to guide the exploration of the reasoning space~\citep{snell2024scalingllmtesttimecompute}. 


Applying test-time scaling effectively requires selecting a suitable decoding strategy (e.g., BoN, Beam Search) and configuring its computational parameters (e.g., number of samples, beam width). Ideally, these choices should reflect the difficulty of the problem: simpler instances can be solved with less compute, while harder ones benefit from greater effort. To support this adaptivity, prior work has attempted to estimate difficulty explicitly, typically by aggregating verifier scores across multiple sampled solutions~\citep{lightman2023letsverifystepstep,snell2024scalingllmtesttimecompute,beeching2024scalingtesttimecompute}. However, this adds significant overhead and yields only a coarse, problem-level estimate. In practice, difficulty often varies \emph{within} a single solution: some steps demand complex reasoning, while others are more straightforward. For instance, when solving a geometry problem, certain steps may require spatial reasoning and the application of a complex theorem, while others might involve simple arithmetic substitutions. Existing methods fail to allocate computation in a way that accounts for this. 

\begin{figure}[t]
    \centering
   \includegraphics[width=0.95\linewidth]
   {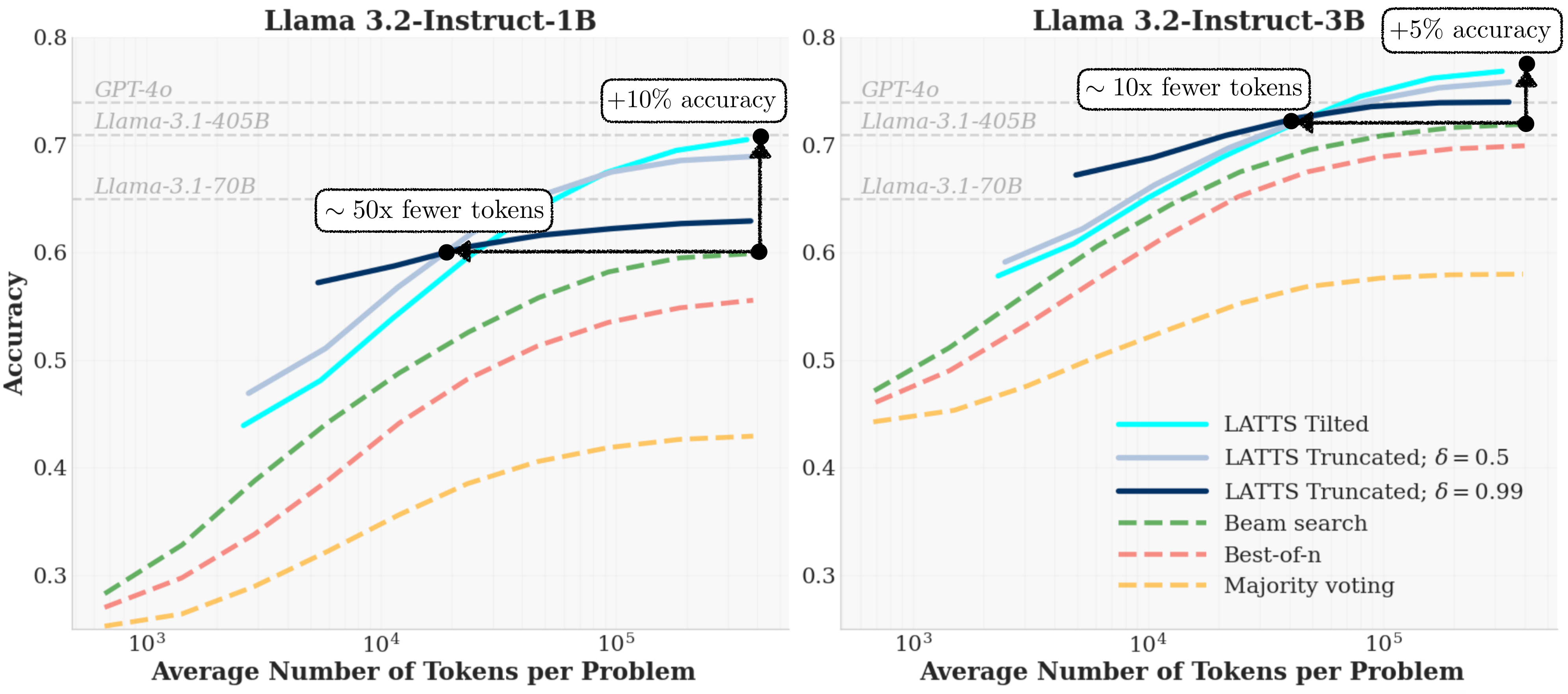}
    \caption{\textbf{Accuracy/compute tradeoff of LATTS vs. standard test-time scaling methods.} LATTS variants perform step-level allocation of computation based on local difficulty, achieving significantly better accuracy/compute tradeoffs across the board. See \S\ref{sec:experiments} for details.}

\label{fig:teaser}
\end{figure}

To address these limitations, we introduce \textit{Locally Adaptive Test-Time Scaling (LATTS)}, a method that dynamically adjusts the amount of computation allocated to each Chain-of-Thought step based on a notion of local difficulty -- without requiring explicit difficulty estimation. LATTS frames the generation process as a sequence of sampling problems from a target distribution defined by the base model modulated by the step-level verifier, relying on an acceptance-rejection scheme to implicitly control computational effort. Our approach is related in spirit to \emph{speculative decoding}~\citep{leviathan2023fastinferencetransformersspeculative,chen2023acceleratinglargelanguagemodel}, which selectively accepts draft tokens as opposed to reasoning steps (see \S~\ref{sec:latts-spec-dec}). Empirical results show that LATTS achieves significantly better accuracy/compute tradeoffs than traditional test-time scaling methods. For example, on MATH500, using Llama 3.2-Instruct-1B~\citep{grattafiori2024llama3herdmodels} as the base model and Qwen2.5-7B~\citep{zhang2025lessonsdevelopingprocessreward} as the verifier, LATTS matches the top performance of Beam Search, the strongest baseline, while generating \textbf{about $\mathbf{50\times}$ fewer tokens}; similarly, using maximal compute budget, LATTS improves Beam Search accuracy \textbf{by more than 10\%} (see Figure~\ref{fig:teaser}).

\vspace{-1mm}
\section{Related Work}
\vspace{-1mm}

\textbf{Chain-of-Thought Reasoning with Step-Level Feedback.}
A growing body of work has started to leverage step-level verifier feedback to improve the performance of Chain-of-Thought (CoT) inference by employing Process Reward Models (PRMs)~\citep{lightman2023letsverifystepstep,zhang2025lessonsdevelopingprocessreward}. In this paradigm, inference is framed as guided exploration through the space of possible reasoning sequences, using verifier scores at each intermediate step to steer the search~\citep{qi2024mutualreasoningmakessmaller,chen2024alphamathzeroprocesssupervision}. A simple implementation is Best-of-N (BoN)~\citep{brown2024large,dong2023raftrewardrankedfinetuning}, which generates multiple reasoning chains concurrently, aggregates their step-level verifier scores (e.g., by taking the minimum or final step score), and selects the highest-scoring candidate. This paper contributes to this line of work by developing algorithms that use verifier scores to adaptively allocate compute to different steps along a CoT-sequence based on the ``difficulty'' of that step.\\[5pt]
\textbf{Adaptive Inference-Time Methods.}
Adaptive inference-time methods aim to improve model performance by dynamically allocating computational resources based on task difficulty~\citep{manvi2024adaptiveinferencetimecomputellms, chen2024magicoremultiagentiterativecoarsetofine}. These approaches differ primarily in how they estimate and act on difficulty signals. For example, \citet{snell2024scalingllmtesttimecompute} estimate problem-level difficulty by averaging PRM scores over multiple sampled solutions, allocating computational resources independently for each problem. In contrast, other methods like~\citep{damani2024learninghardthinkinputadaptive,wang2025makepennycountdifficultyadaptive} operate at a batch level: given a set of problems and a fixed computational budget, they predict relative difficulties within a batch and distribute resources accordingly. However, none of these methods address local variations in difficulty that arise within individual reasoning chains for a single problem,e.g., when only certain steps in a solution require substantial effort. Also related are speculative decoding methods~\citep{leviathan2023fastinferencetransformersspeculative,chen2023acceleratinglargelanguagemodel}, which accelerate generation by sampling from a smaller draft model and verifying outputs with a stronger model. While typically applied at the token level, they share conceptual similarities with our use of verifier-guided sampling to control compute. Most closely related is the recently proposed ``reward-guided'' speculative decoding~\citep{liao2025rewardguidedspeculativedecodingefficient}, which uses PRM feedback to selectively invoke a larger model at difficult reasoning steps. In contrast, LATTS operates entirely within a single smaller model, using verifier feedback to repeatedly sample challenging steps and recover from failures through strategies like backtracking or restarting. We discuss these differences in more detail in \S\ref{sec:latts-spec-dec}.

\section{Setup}
\label{sec:setup}

We consider the general setting of problem-solving with large language models (LLMs), where a model is given a natural language input (e.g., a math question), and the goal is to generate a free-text output from which a final answer can be extracted. In this work, we focus on settings where solutions are expressed as a sequence of intermediate reasoning steps.

\paragraph{Language model.} Let \(\mathcal{A}\) be a finite vocabulary, and \(\mathcal{A}^*\) the set of finite token sequences. A language model defines a distribution \(p_{\text{model}} \in \mathcal{P}(\mathcal{A}^*)\) over strings. We structure generation as a sequence of reasoning steps \(S = s_1, \dots, s_T\), where each step \(s_t \in \mathcal{A}^*\) is a coherent textual span, and the final step \(s_T\) contains the answer. Generation is modeled auto-regressively at the step level, that is,
$p_{\text{model}}(S \mid x) = \prod_{t=1}^{T} p_{\text{model}}(s_t \mid x, s_1, \dots, s_{t-1}).$ We encourage such structured reasoning using Chain-of-Thought (CoT) prompting~\citep{wei2023chainofthoughtpromptingelicitsreasoning}, where individual steps are separated by a special delimiter -- in our setting, a double newline ``\texttt{\textbackslash n\textbackslash n}''.


\paragraph{Compute-accuracy tradeoff.} To evaluate the effectiveness of different test-time scaling strategies, we consider their \emph{compute-accuracy tradeoff}. Specifically, we compare methods based on the accuracy they achieve as a function of average computational cost. A method is said to be more effective if its tradeoff curve \emph{dominates} another: that is, for any fixed compute budget, it achieves equal or better accuracy, or equivalently, achieves a target accuracy with lower compute. In this work, we mostly define computational cost as the total number of generated tokens (including any rejected candidates). This is a practical metric for real-world applications where cost is tied to token usage (e.g., when using commercial APIs). While token-based compute is not always ideal—for example, in scenarios with batch decoding or parallel sampling—we focus on this measure for concreteness and revisit batching in later sections. 

\paragraph{Verifier.} We assume access to a \emph{verifier} function \(r: \mathcal{A}^* \times \mathcal{A}^* \to [0,1]\) that assigns a (continuous) scalar score to a reasoning step \(s_t\), given the input \(x\) and the previously generated steps \(S_{<t} = s_1, \dots, s_{t-1}\). The score \(r(s_t \mid x, S_{<t})\) reflects how likely \(s_t\) is to be correct or helpful in progressing toward a valid final answer. In practice, \(r\) may be implemented using a process reward model (PRM)~\citep{lightman2023letsverifystepstep, snell2024scalingllmtesttimecompute}, or by prompting a stronger LLM to act as a critic~\citep{zheng2023judgingllmasajudgemtbenchchatbot, ankner2024critiqueoutloudrewardmodels}, e.g., with a question such as \texttt{Is this step correct?~Yes/No} and using the predicted confidence in \texttt{Yes} as the verifier score.


\paragraph{Modulator.} To control the selectivity of the verifier, we can use a \emph{modulator} function \(f: [0,1] \to [0,1]\), where \(f\) is non-decreasing and satisfies \(f(0) = 0\), \(f(1) = 1\). The \emph{modulated verifier} is defined as \(r_f = f \circ r\), and allows us to vary the stringency of step acceptance. Two natural choices are:
\begin{itemize}[leftmargin=1.2em,itemsep=2pt]
\item \textit{Identity:} \(f(z) = z\), which uses the raw verifier score.
\item \textit{Threshold:} \(f(z) = \mathbf{1}\{z \geq \delta\}\), which applies a hard threshold at \(\delta\). This allows for a more interpretable notion of ``correctness'' of a reasoning step by casting it as a binary decision.
\end{itemize}


\section{Method}
\label{sec:method}
Our approach dynamically allocates computational effort across reasoning steps, using a verifier to decide whether to accept, reject, or retry a step. The core mechanism is based on acceptance-rejection (AR) sampling so that difficult steps automatically require more trials before acceptance. If no valid step is found within a fixed number of trials, we invoke a fallback mechanism, such as terminating early, choosing the best-failed candidate, backtracking to earlier steps, or restarting the entire generation. We describe each component in detail below.

\subsection{Verifier-Guided Acceptance-Rejection Sampling}
\label{sec:ar_sampling}
Let \(p_{\text{model}}(\cdot \mid x, S_{<t})\) be the language model's distribution over the next reasoning step, given the input \(x\) and current partial chain \(S_{<t} = s_1, \dots, s_{t-1}\). Let \(r_f(s_t \mid x, S_{<t})\) be a modulated verifier score for a candidate step \(s_t\), as described in Section~\ref{sec:setup}. We define a target distribution over reasoning steps that assigns a higher likelihood to steps that are (i) more linguistically coherent, and (ii) more likely to help solve the question (higher verifier correctness):
\begin{equation}
p_{\text{target}}(s_t \mid x, S_{<t}) \propto \underbrace{p_{\text{model}}(s_t \mid x, S_{<t})}_{\textrm{Linguistic coherence}} \cdot \underbrace{r_f(s_t \mid x, S_{<t})}_{\textrm{Verifier correctness}}.
\end{equation}

Direct sampling from $p_{\text{target}}$ is unfortunately intractable, so we use acceptance-rejection (AR) sampling \citep{robert1999monte} with \(p_{\text{model}}\) as the proposal and \(r_f\) as the acceptance probability. Specifically, for each reasoning step:
\begin{enumerate}[leftmargin=1cm,itemsep=0.1cm,topsep=0.1cm,partopsep=0cm,parsep=0.1cm]
\item Sample candidate step \(s_t \sim p_{\text{model}}(\cdot \mid x, S_{<t})\);
\item Sample \(u \sim \mathcal{U}([0,1])\);
\item Accept \(s_t\) if \(u < r_f(s_t \mid x, S_{<t})\), otherwise reject and repeat.
\end{enumerate}
This procedure guarantees that the accepted samples are drawn from \(p_{\text{target}}\). However, if no candidate is accepted within a fixed number of trials, we invoke a fallback strategy such as backtracking or early termination (see \S\ref{sec:handling_rejection}). Following the above procedure, generation proceeds step-by-step until a stopping condition is met (for example, a final answer is detected or a maximum number of steps is reached). Our general approach, termed \textit{Locally Adaptive Test-Time Scaling (LATTS)}, consists in sequentially generating each reasoning step using this AR scheme. 


\begin{figure}[t]
    \centering
   \includegraphics[width=0.85\linewidth]
   {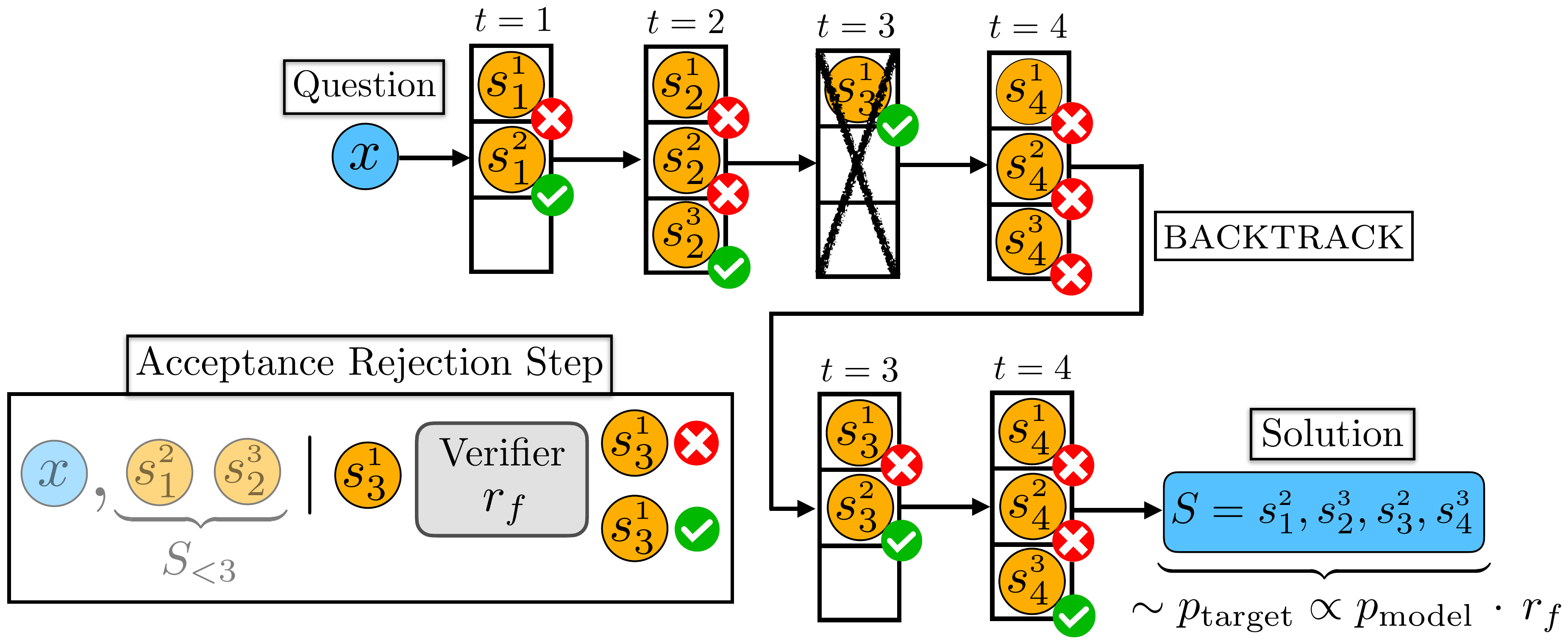}
\caption{\textbf{Illustration of LATTS} with max trials per step \(M=3\) and \textsc{Backtrack} fallback. Conditioned on \(x\), reasoning steps are generated via acceptance-rejection sampling using proposal \(p_\text{model}\) and target \(p_\text{target} \propto p_\text{model} \cdot r_f\), where \(r_f = f \circ r\). Step 1 accepts \(s_{1}^{2}\); Step 2 accepts \(s_{2}^{3}\) after two rejections. Step 3 accepts \(s_{3}^{1}\), but Step 4 rejects all candidates, triggering a backtrack: step 3 is redone, accepting \(s_{3}^{2}\). Step 4 then accepts \(s_{4}^{3}\), yielding the final solution \(S = s_{1}^{2}, s_{2}^{3}, s_{3}^{2}, s_{4}^{3}\).}
\label{fig:latts_concept_figure}
\end{figure}

\subsection{Variants of LATTS via Modulated Verifiers}
\label{sec:different-targets}

Different choices of the modulator function \(f\) lead to different instantiations of LATTS, each corresponding to a distinct target distribution. 
\begin{enumerate}
    \item \textbf{LATTS Tilted:} When \( f(z) = z \), the target corresponds to the model \emph{tilted} by $r$:
    \begin{equation}
    \label{eq. latts tilted eq}
    p_{\text{target}}(\cdot \mid x, S_{<t}) \propto p_{\text{model}}(\cdot \mid x, S_{<t}) \cdot r(\cdot \mid x, S_{<t}),
    \end{equation}
    Under this choice, the AR procedure repeatedly samples until it finds a step whose score exceeds a stochastically sampled threshold, specifically \(r \geq u\) with \(u \sim \mathcal{U}[0, 1]\).

    \item \textbf{LATTS Truncated:} When \( f(z) = \mathbf{1}\{z \geq \delta\} \), the target corresponds to the model \emph{truncated} by $r$:
    \begin{equation}
    \label{eq. latts truncated eq}
    p_{\text{target}}(\cdot \mid x, S_{<t}) = p_{\text{model}}(\cdot \mid x, S_{<t}, r \geq \delta).
    \end{equation}
    See Appendix \S\ref{subsec:simplication-truncated}. Under this choice, the AR procedure repeatedly samples until finding a step with a score \(r \ge \delta\). 
\end{enumerate}

\textbf{Connections with Reinforcement Learning}. Interestingly, the target distributions in \eqref{eq. latts tilted eq} and \eqref{eq. latts truncated eq} correspond to the optimal policy when solving a KL-regularized reward maximization problem \citep{schaal2007, korbak2022reinforcementlearningdistributionmatching,go2023aligninglanguagemodelspreferences,rafailov2024directpreferenceoptimizationlanguage} Formally, 
\begin{equation}
    p_{\text{target}}(\cdot \mid x, S_{<t}) = \argmax_{p \,\in\, \mathcal{P}(\mathcal{A}^*)} \mathbb{E}_{s_t \sim p}[r_f(s_t \mid x, S_{<t})] - \text{KL}(p \mid p_{\text{model}}(\cdot \mid x, S_{<t})).
\end{equation}



\subsection{Local Difficulty and Implicit Effort Modulation}
\label{sec:local-difficulty}

The AR sampling procedure provides an implicit mechanism for adaptively scaling computation: steps that are more ''difficult'' require more sampling attempts before an acceptable candidate is found. We can formalize this notion of step-level difficulty as follows:

\begin{definition}
Let \(x\) be a problem, \(S_{<t}\) a partial reasoning chain, and \(r_f\) a modulated verifier. The \emph{local difficulty} of step \(t\) is defined as:
\begin{equation}
\label{eq:local-difficulty}
\Delta(x, S_{<t}) = 1 - \mathbb{E}_{s_t \sim p_{\text{model}}(\cdot \mid x, S_{<t})}\left[ r_f(s_t \mid x, S_{<t}) \right].
\end{equation}
\end{definition}

This quantity lies in $[0, 1]$, with higher values indicating greater difficulty. When $\Delta$ approaches~$1$, steps sampled from $p_{\text{model}}$ tend to receive low verifier scores, indicating the model struggles to generate steps the verifier considers correct. Conversely, when $\Delta$ approaches $0$, the model consistently generates steps that receive high verifier scores. Note that our notion of local difficulty is relative to the model employed to solve the task. 

    

\textbf{Connection with AR sampling scheme}.
For any step $t$, let $n_t$ denote the number of trials needed by the AR sampler before accepting a candidate reasoning step $s_t$. Then, the expected number of trials needed by our AR sampler to accept a candidate is given by
\begin{equation}
\label{eq: conection diff. and ar}
    \mathbb{E}[n_t] = (\mathbb{E}_{s_t \sim p_{\text{model}}(\cdot \mid x, S_{<t})}[r_f(s_t \mid x, S_{<t})])^{-1} = \frac{1}{1 - \Delta(x, S_{<t})}.
\end{equation}
 This follows from standard results, but we include a proof in Appendix \S\ref{subsec:proof ar local difficulty}. 
Thus, in expectation, our AR sampler allocates more compute to steps with higher local difficulty and vice-versa. 

\subsection{Handling Rejection: Fallback Strategies}
\label{sec:handling_rejection}
In practice, it is computationally infeasible to draw from \(p_{\text{target}}\) using an unbounded number of AR trials. We therefore impose a maximum number of candidate attempts per step, denoted by \(M\). If no candidate is accepted within \(M\) trials, we apply a fallback \textsc{action} to maintain progress. We consider four fallback strategies:
\begin{itemize}[leftmargin=0.5cm,itemsep=1pt]
    \item \textsc{stop}: Terminate the chain and return a default fallback answer (e.g., ''I don't know'').
    \item \textsc{max}: Choose the candidate step with the highest verifier score among the rejected samples, i.e., $s_t^{\text{max}} = \arg\max_{m =1,\ldots,M} r(s_t^m \mid x, S_{<t}),$
    and continue generation with this step.
    \item \textsc{backtrack}: Remove the previous step \(s_{t-1}\), revert to \(S_{<t-1}\), and reattempt step \(t-1\) using fresh candidates (see Figure~\ref{fig:latts_concept_figure} for an illustration).
    
    \item \textsc{restart}: Clear the full chain \(S_{<t}\), and restart generation from scratch at \(t = 1\).
\end{itemize}


To prevent infinite generation loops, we introduce a maximum number of fallback actions \(A\). We study the effectiveness of each fallback \textsc{action} and analyze the impact of \(A\) in Figure~\ref{fig:merged} (right). The complete LATTS procedure, including our fallback strategy, is summarized in Algorithm~\ref{algo:latts}. Additionally, we provide an illustrative example of LATTS with the \(\textsc{backtrack}\) fallback in Figure~\ref{fig:latts_concept_figure}. We additionally provide  illustrative example of LATTS for the other fallbacks in Appendix~\ref{sec:additional-figures}, as well as reasoning chains produced by LATTS on actual math problems in Appendix~\ref{sec:reasoning-traces}.

\subsection{Aggregating Multiple LATTS Completions}
\label{sec: agg. latts}

As in standard test-time scaling, we can improve performance by generating \(N\) independent LATTS completions and selecting a final answer using a voting strategy. Each completion includes a verifier score for its final step, which we use to weigh votes. Instead of selecting the highest-scoring completion directly, we follow prior work~\citep{beeching2024scalingtesttimecompute,snell2024scalingllmtesttimecompute} and perform weighted majority voting: we map each final answer to a canonical form and select the one with the highest total verifier score across all completions. This combines LATTS's refined sampling with the benefits of parallel generation.


\subsection{LATTS as Speculative Decoding over Reasoning Steps}
\label{sec:latts-spec-dec}

LATTS can be interpreted as a form of speculative decoding~\citep{leviathan2023fastinferencetransformersspeculative, chen2023acceleratinglargelanguagemodel}, adapted for multi-step free-text reasoning. While traditional speculative decoding aims to accelerate generation by validating token-level drafts from a fast model using a stronger model, LATTS reinterprets this paradigm at the level of reasoning steps. In our formulation:
\begin{itemize}[leftmargin=1.2em,itemsep=0.3pt]
    \item The draft model is the given language model \(p_{\text{model}}\);
    \item The target model is implicitly defined by \(p_{\text{target}} \propto p_{\text{model}} \cdot r_f\);
    \item The verification takes place at the level of reasoning steps rather than individual tokens.
\end{itemize}

Unlike traditional speculative decoding, where the target model is accessible and can regenerate rejected tokens, our target is only defined implicitly. Moreover, it operates over a combinatorially large space of free, text strings, rendering enumeration infeasible. As a result, resampling under LATTS relies solely on the draft model to propose alternative steps.

Concurrently with our work, \citet{liao2025rewardguidedspeculativedecodingefficient} propose a reward-guided speculative decoding method that also employs step-level verifier feedback. However, their approach assumes access to two distinct generative models: a draft model and a stronger target model. When a proposed step is rejected, they fall back to sampling from the stronger model, resulting in a final distribution that is effectively a mixture over both models, conditioned on the verifier’s accept/reject decisions. In contrast, LATTS does not require two generative models, and directly targets the distribution \(p_{\text{target}} \propto p_{\text{model}} \cdot r_f\).

\section{Experiments}
\label{sec:experiments}


All experiments were conducted on 8 NVIDIA A100 GPUs, using vLLM for inference serving~\citep{kwon2023efficientmemorymanagementlarge}. Below, we detail the models, datasets, and baselines used in our experiments.

\begin{figure}[t]
    \centering
    \includegraphics[width=0.95\linewidth]{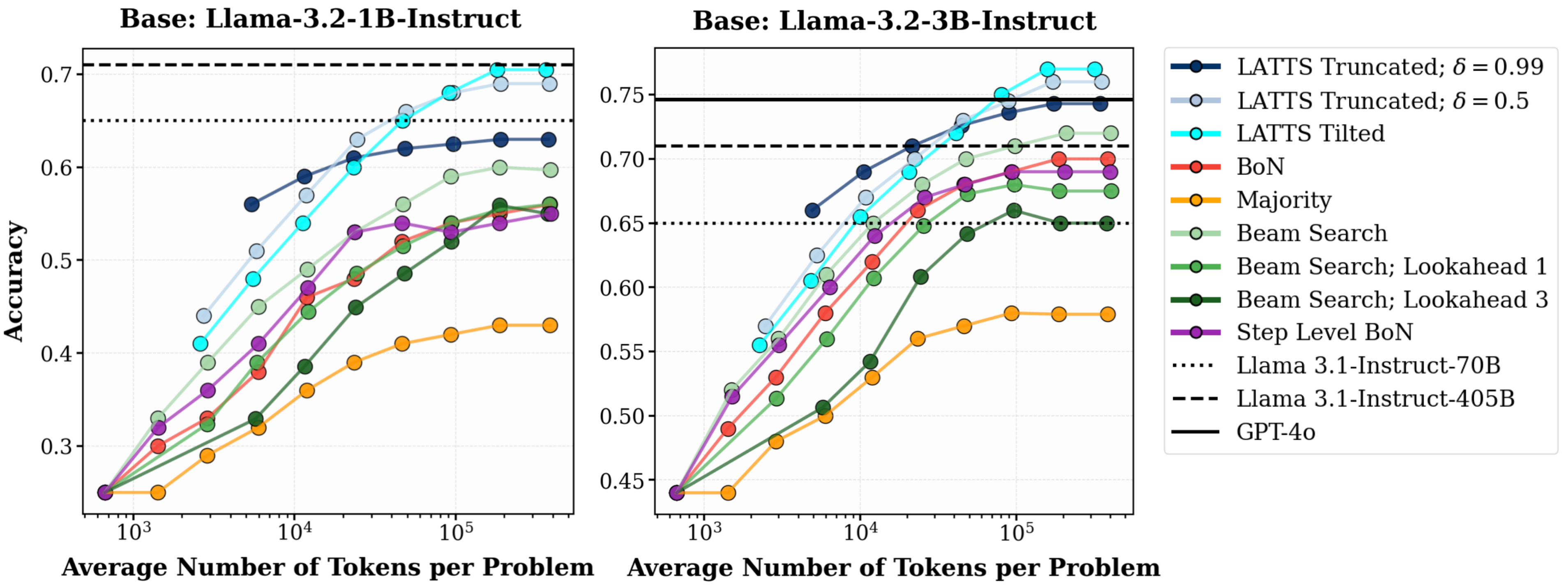}
    \caption{\textbf{Results on MATH500 using PRM as verifier.} Accuracy vs. token usage for different LATTS variants and standard test-time scaling baselines. Each point in the plot corresponds to a particular number of completions used. For reference, we add zero-shot performance of larger models, namely Llama-3.1-Instruct 70B,  Llama-3.1-Instruct 405B, and GPT-4o. LATTS consistently dominates across the compute spectrum.}
    \label{fig:prm}
\end{figure}

\textbf{Models.} We use Llama 3.2-Instruct-1B and 3B~\citep{grattafiori2024llama3herdmodels} as base models (i.e., our $p_{\textrm{model}}$). For verification, we primarily use Qwen2.5-7B PRM~\citep{zhang2025lessonsdevelopingprocessreward}, shown in prior work to be one of the strongest open-source PRM models~\citep{zheng2024processbenchidentifyingprocesserrors,he2025largelanguagemodelsdetect}. We also experiment with prompting Llama 3.1-Instruct-70B as a critic. \\[2.5pt]
\textbf{Datasets.} We evaluate on two math benchmarks: (1) MATH500, a 500-example subset of the MATH test set~\citep{lightman2023letsverifystepstep}; (2) AIME~\citep{aimo_validation_aime}, a set of 90 challenging Olympiad-level problems sourced from competitions held between 2022 and 2024.\\[2.5pt]
\textbf{Generation settings.} We use nucleus sampling ($p = 0.9$) \citep{holtzman2019curious} with temperature $\tau = 0.8$, and delimit reasoning steps with double newlines. LATTS is run with up to $M=32$ trials per step and fallback limit $A=8$ unless otherwise noted.\\[2.5pt]
\textbf{Baselines.} We compare against common test-time scaling methods: Majority Voting, BoN, step-level BoN, Beam Search, and Beam Search with lookahead. See Appendix~\ref{sec:tts-reminder} for details.\\[2.5pt]
\textbf{Evaluation metric.} We report accuracy (fraction of correctly solved problems) as a function of average number of generated tokens per problem. This reflects both quality and cost efficiency. In Appendix~\S\ref{sec:chunkwise-trials} (see notably Figure~\ref{fig:chunk-wise-generation}), we investigate how batching the generation of trials can reduce answer latency at the expense of generating additional tokens. 

Although verifier-based test-time scaling methods typically incur an additional cost of calling the verifier, for simplicity we do not consider this cost for our accuracy-compute tradeoffs. Nevertheless, we note that LATTS substantially reduces the number of these calls (per problem) compared to our strongest baseline, Beam Search.\footnote{For context, on MATH500, LATTS truncated with $\delta = 0.99$ results in $\sim 20$ verifier calls per problem on average. In comparison, our best performing Beam Search baseline, with 4 beams and beam width of 2, requires $\sim 80$ verifier calls per problem --- a $4\times$ increase compared to LATTS truncated despite generating a similar number of tokens.}



\begin{figure}[t]
    \centering
   \includegraphics[width=0.95\linewidth]
   {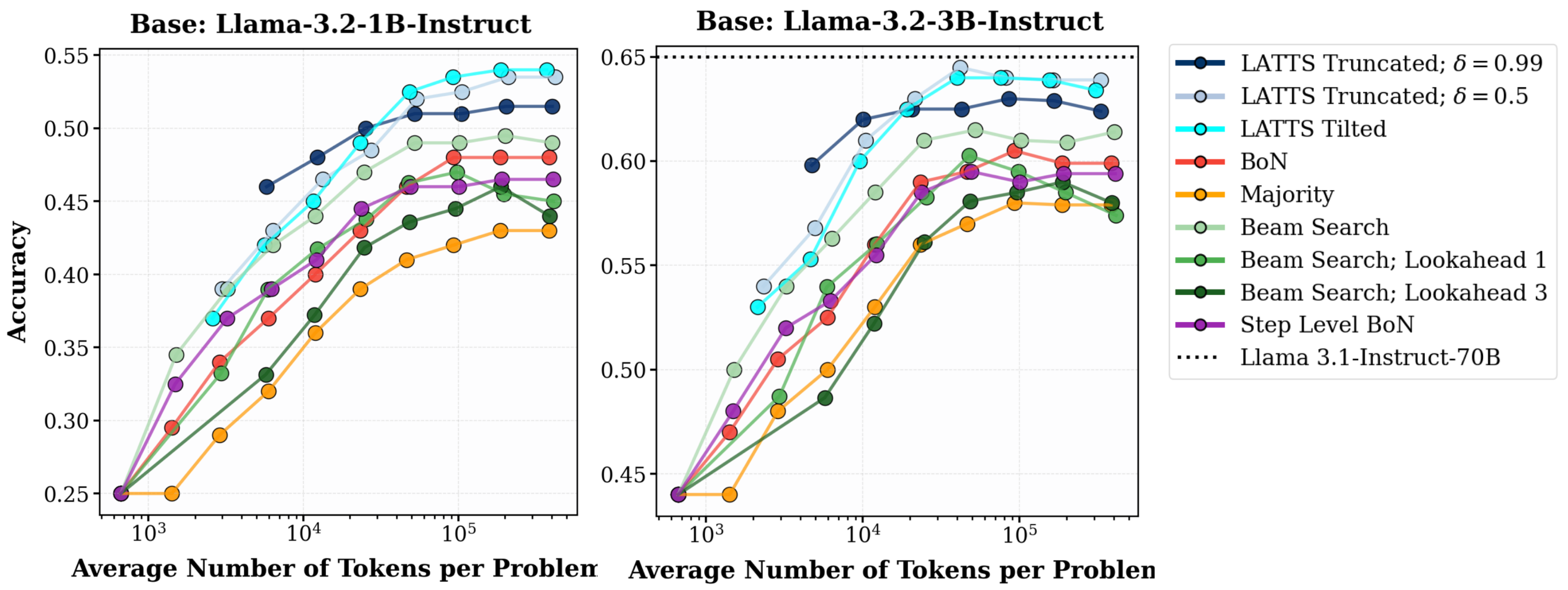}
    \caption{\textbf{Result on MATH500 using LLM Critic as a Verifier.} Here we use Llama 3.1-70B as our LLM critic. Each point in the plot corresponds to a particular number of completion used.} 
\label{fig:llm_as_critique}
\end{figure}

\subsection{Results}

\paragraph{MATH500 with PRM verifier.}

We evaluate LATTS on the MATH500 benchmark using Qwen2.5-7B PRM as the verifier and compare against standard test-time scaling baselines. Figure~\ref{fig:prm} reports accuracy as a function of total generated tokens. Here, we scale compute by increasing the number of completions \(N\) (See \S\ref{sec: agg. latts}), ranging from 1 to 32 for LATTS, and up to 128 for baselines like Majority Voting and BoN.
Across all settings, LATTS methods consistently outperform baselines, achieving strictly better accuracy–compute tradeoffs. For example, using Llama 3.2-Instruct-1B as the base model, with a fixed compute budget of $10^5$ generated tokens, LATTS Tilted outperforms the strongest baseline (Beam Search) by about 15\%. Similarly, for a fixed target accuracy of $0.5$, LATTS Tilted requires about $10\times$ fewer tokens per problem on average. 
Interestingly, no single LATTS variant dominates in absolute terms across the full compute range. At low budgets (e.g., \(N=1\)), LATTS Truncated with \(\delta = 0.99\) yields the highest accuracy; at higher budgets, LATTS Tilted and LATTS Truncated with \(\delta = 0.5\) are more effective. We further analyze this effect later on in \S\ref{sec. ablations}.

Remarkably, scaling test-time compute using LATTS allows for base models to be competitive with significantly larger LLMs. For example, using LATTS Tilted with \(N=32\) completions, a Llama-3.2-1B-Instruct model can match the performance of Llama 3.1-Instruct-405B and exceed Llama 3.1-Instruct-70B by 5.1\% on MATH500 (see Figure \ref{fig:prm}). Similarly, with the 3B model, LATTS Tilted surpasses Llama-70B by 10.8\%, Llama-405B by 4.6\%, and even outperforms GPT-4o by 1.6\%. 

\paragraph{MATH500 with LLM Critic as Verifier.}

\begin{wrapfigure}{o}{0.55\textwidth}
    \includegraphics[width=\linewidth]{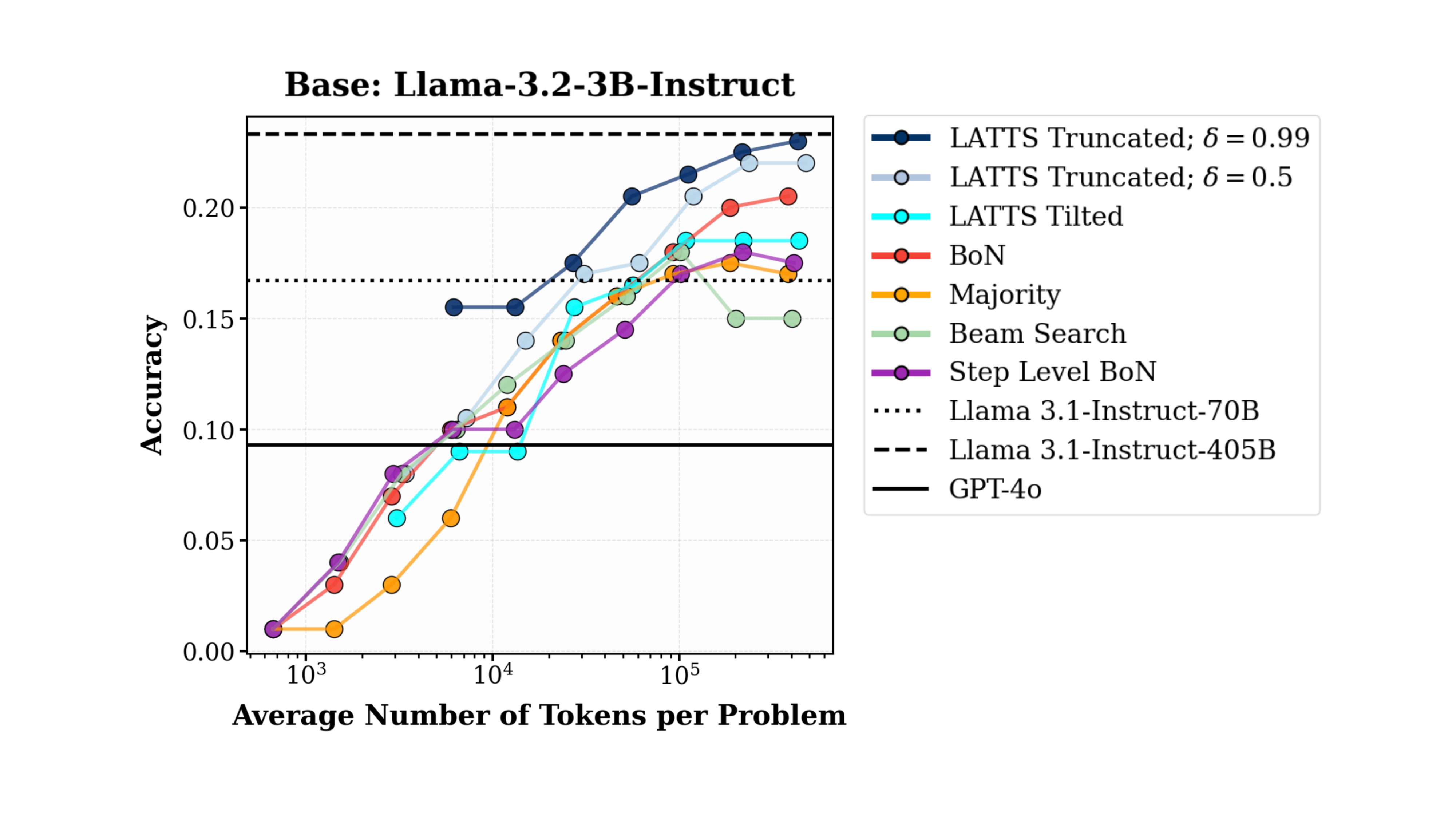}
    \caption{\textbf{Results on AIME using PRM as verifier.} Each point in the plot corresponds to a particular number of completions used.}
    \label{fig:aime}
\end{wrapfigure}

Learning a dedicated Process Reward Model (PRM) for a particular task (e.g., solving math questions) requires large-scale data annotation which is often not available \textit{a priori} and is challenging to curate \citep{yuan2024free}. 

Consequently, we explore the possibility of using an LLM as a verifier to score reasoning steps for LATTS, bypassing the need for a dedicated PRM. Specifically, we prompt Llama 3.1-Instruct-70B as a step-level critic: Given the original question \(x\), the reasoning prefix \(S_{<t}\), and a candidate step \(s_t\), we ask \texttt{ `Is this step correct? Yes/No'}. The verifier score is then computed as the softmax probability assigned to the \texttt{Yes} token, providing a natural confidence score in the range \([0,1]\).

Figure~\ref{fig:llm_as_critique} shows that all LATTS variants continue to dominate standard baselines under this setup. Although the overall accuracy is slightly lower compared to the use of Qwen2.5-7B PRM --- consistent with previous observations that general-purpose LLMs are less reliable as step-level verifiers~\citep{zheng2024processbenchidentifyingprocesserrors}---the relative trends remain remarkably consistent. LATTS Truncated with \(\delta = 0.99\) achieves the best performance at low compute, while LATTS Tilted becomes most effective as the number of completions increases.


\begin{figure}[t]
    \centering
   \includegraphics[width=0.95\linewidth]
   {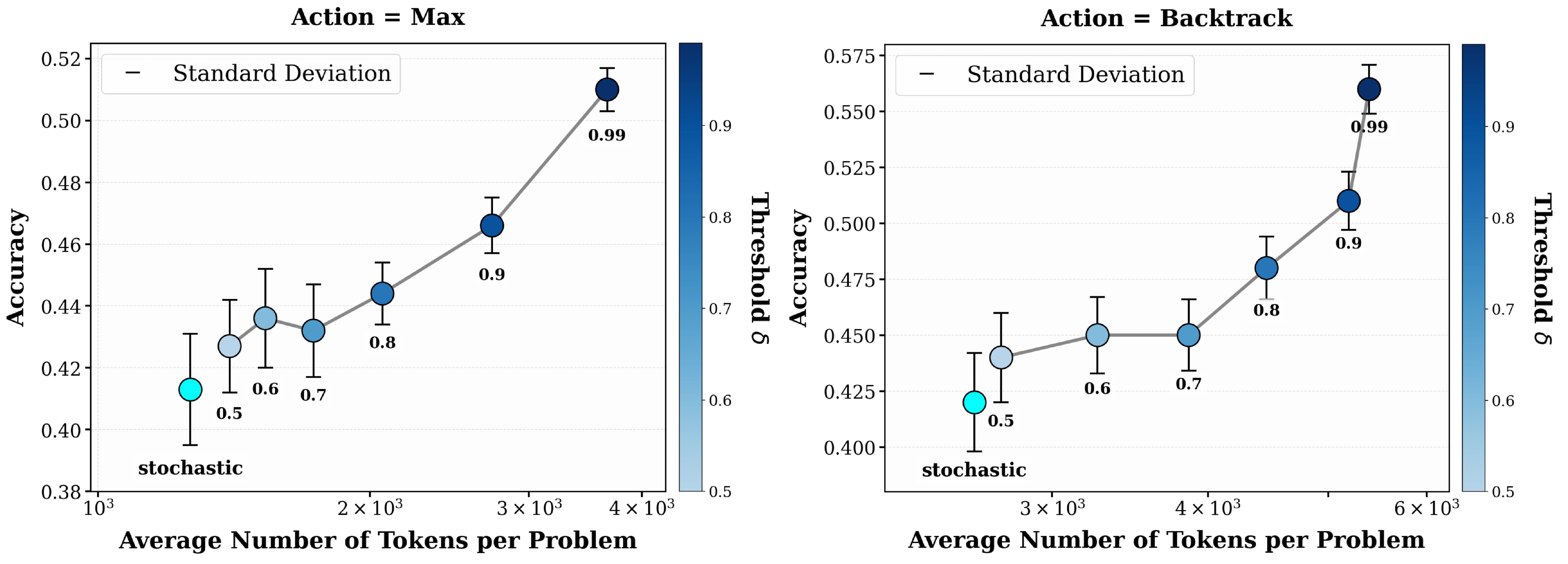}
    \caption{\textbf{Effect of the modulator on single-completion accuracy.}  We use Llama 3.2-Instruct-1B as the base model and Qwen2.5-7B PRM as the verifier, averaging accuracy over 10 single-completion runs per problem on the MATH500 dataset. We vary the modulator by using \( f(z) = \mathbf{1}\{z \geq \delta\} \) with thresholds \(\delta \in \{0.5, 0.6, \ldots, 0.9, 0.99\}\) (yielding deterministic thresholding), or (\(f(z)=z\)) (yielding stochastic thresholding with $0.5$ expectation). Two fallback strategies,\textsc{max} and \textsc{backtrack}, are considered when all candidate steps are rejected.}
\label{fig:threshold}
\end{figure}

\paragraph{AIME with PRM verifier.}
We also evaluate LATTS on the AIME dataset, which consists of 90 Olympiad-level problems that are significantly more challenging than those in MATH500. All experiments here use Llama 3.2-Instruct-3B as the base model and Qwen2.5-7B PRM as the verifier.

As shown in Figure~\ref{fig:aime}, LATTS Truncated with \(\delta = 0.99\) consistently dominates all baselines across the full compute range (beam search with lookahead is omitted due to poor performance). At a high budget (large number of completions), this LATTS variant matches the performance of the much larger Llama 3.1-Instruct-405B model.

Unlike in MATH500, where softer modulators (e.g., Tilted or \(\delta = 0.5\)) perform better at high compute, the strictest variant (\(\delta = 0.99\)) remains the most effective throughout.  This suggests that challenging problems, increasing diversity through relaxed step acceptance criterion introduces more failure modes, as even small reasoning errors can undermine the final answer.

\subsection{Ablations}
\label{sec. ablations}

\paragraph{Effect of the Modulator on Single-Completion Performance.}
We analyze how the choice of modulator \(f\) affects single-completion accuracy ($N=1$).  Figure~\ref{fig:threshold} shows accuracy on MATH500 using Llama 3.2-Instruct-1B with the PRM verifier. For the hard modulator \(f(z) = \mathbf{1}\{z \geq \delta\}\), increasing the threshold \(\delta\) leads to steady improvements in accuracy.  These gains come at the cost of increased computation, as more candidate steps are rejected. Overall, these results show that adjusting the threshold in the hard modulator offers an additional axis for scaling test-time computation. Higher thresholds lead to better single-completion accuracy by investing more effort in difficult steps. This complements the more typical form of test-time scaling via multiple completions and allows fine-grained control over both cost and reliability.

\paragraph{Effect of the Fallback Strategy.}

\begin{figure}[t]
    \centering
   \includegraphics[width=0.95\linewidth]
   {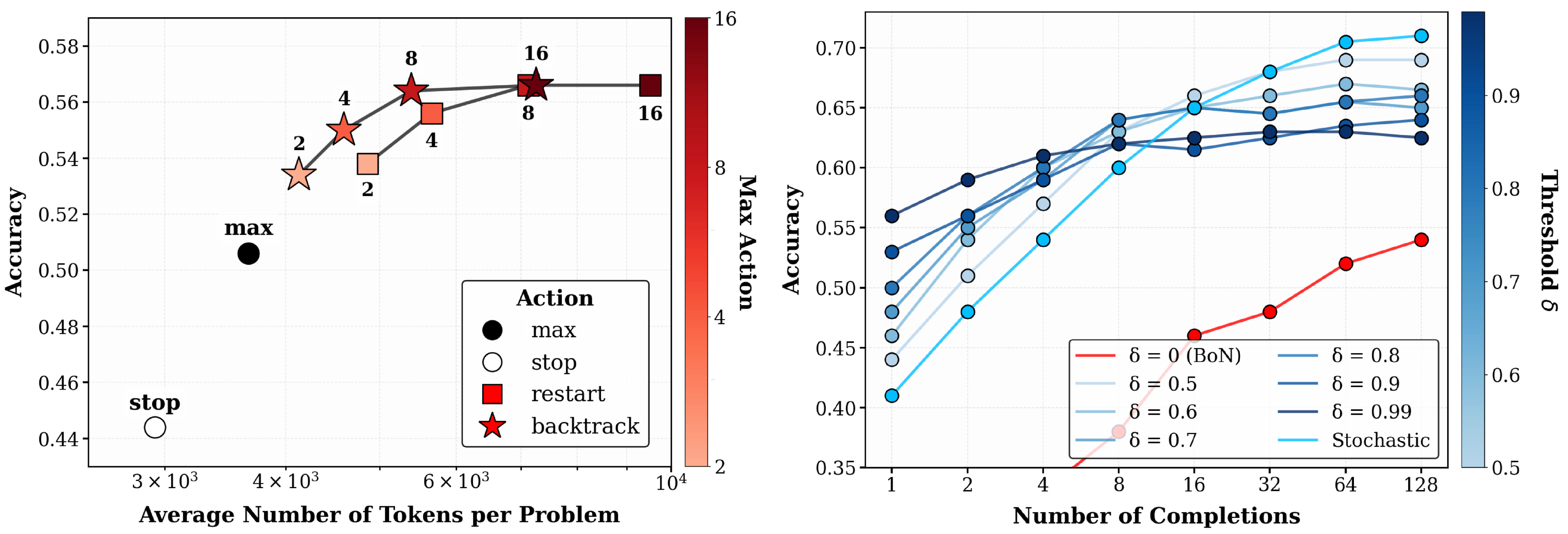}
 \caption{\textbf{Fallback strategies (left) and modulator effects on multi-completion accuracy (right).} \textbf{Left:} We compare fallback strategies under a strict modulator (\(\delta = 0.99\)) on MATH500 with single completions. For \textsc{Restart} and \textsc{Backtrack}, we vary the maximum number of allowed fallback actions (as indicated by numbers on each point in the plots). Overall, \textsc{Backtrack} achieves the best tradeoff between accuracy and cost. \textbf{Right:} We analyze how modulator choice affects accuracy when aggregating multiple completions. Lower thresholds increase diversity and improve aggregated accuracy, with Tilted (\(f(z) = z\)) achieving the highest performance at \(N = 128\). }
\label{fig:merged}
\end{figure}

We next examine how different fallback strategies affect performance when no step is accepted after the maximum number of trials (\(M = 32\)). Figure~\ref{fig:merged} (left) compares four strategies, \textsc{stop}, \textsc{max}, \textsc{backtrack}, and \textsc{restart}. For \textsc{backtrack} and \textsc{restart}, we vary the maximum number of allowed fallback actions \(A\), while the other strategies do not use this parameter. We evaluate all methods using LATTS Truncated with \(\delta = 0.99\) that frequently triggers rejections. \textsc{Backtrack} consistently achieves higher accuracy than \textsc{restart}, while also incurring lower computational cost. In contrast, \textsc{stop} and \textsc{max} are more efficient but perform worse.

\paragraph{Effect of the Modulator on Multi-Completion Performance.}

We finally study how the choice of modulator affects performance when aggregating multiple LATTS completions using weighted majority voting. As shown in Figure~\ref{fig:merged} (right), the best-performing modulator depends on the number of completions. For \(N = 1\), high thresholds (e.g., \(\delta = 0.99\)) yield the most accurate individual completions. As \(N\) increases, this trend reverses, with lower thresholds producing significantly better aggregated results. These results show that the modulator threshold behaves similarly to a temperature parameter: high thresholds yield low-diversity, high-confidence outputs, akin to low-temperature decoding, while low thresholds increase variability at the cost of occasional errors. We also observe that the BoN baseline (equivalent to \(\delta = 0\)) underperforms even at high \(N\), indicating that while some diversity improves aggregation, excessive diversity without filtering can degrade answer quality. 
\section{Conclusion}

In this paper, we introduced Locally Adaptive Test-Time Scaling (LATTS), an approach that dynamically allocates computational resources during language model inference. Through extensive empirical evaluation, we demonstrated that LATTS achieves comparable or superior performance to existing verifier-based methods while using significantly fewer tokens, up to $100\times$ fewer in mathematical reasoning tasks. Our method proved effective across different base models, verifier configurations (including both dedicated PRMs and black-box critique models), and problem difficulties (from standard math problems to olympiad-level challenges). Moreover, the efficacy of our method using LLMs as verifiers opens the possibility of extending LATTS to other domains (beyond math) where trained PRMs are scarcely available.



\appendix
\section*{Appendix}
\label{sec:appendix}

\section{Common Test-Time Scaling Strategies}
\label{sec:tts-reminder}
In this section, we review common test-time scaling strategies.

\paragraph{Majority Voting.}
Given a fixed number of completions \(N\) per question \(x\), we independently generate reasoning chains \(S^1,\dots,S^N\). We extract the final answers \(s^1_{T_1},\dots,s^N_{T_N}\), convert them into canonical mathematical forms \(a^1,\dots,a^N\), and select the most frequent canonical answer:
\[
a^\star = \arg\max_{a^i}\sum_{j=1}^{N}\mathbf{1}\{a^j=a^i\}.
\]
\paragraph{Best-of-N (BoN).}
BoN corresponds to a weighted majority voting scheme. Similar to Majority Voting, BoN weights each canonical answer by the verifier score of the corresponding final reasoning step:
\[
a^\star = \arg\max_{a^i}\sum_{j=1}^{N}r(s^j_{T_j}\mid x, S^j_{<T_j})\mathbf{1}\{a^j=a^i\}.
\]

\paragraph{Step-Level BoN.}
Rather than evaluating reasoning chains only at the end, Step-Level BoN evaluates each reasoning step individually through the following iterative procedure:
\begin{enumerate}
    \item \textbf{Generation:} Generate \(N\) candidate steps \(s_i^1,\dots,s_i^N\). 
    \item \textbf{Scoring:} Evaluate each candidate step using verifier scores:
    \[
    r(s_i^1\mid x,s_1\dots s_{i-1}),\dots,r(s_i^N\mid x,s_1\dots s_{i-1}).
    \]
    \item \textbf{Selection:} Select the highest-scoring candidate as the next reasoning step:
    \[
    s_{i}^\star = \arg\max_{s_i^j} r(s_i^j\mid x,s_1\dots s_{i-1}).
    \]
\end{enumerate}
The final answer is obtained through weighted majority voting as in BoN.

\paragraph{Beam Search.}
Beam Search optimizes the verifier’s predictions through systematic exploration of its step-wise estimates. Given an initial number of beams \(N\) and a beam width \(M\), Beam Search proceeds iteratively:
\begin{enumerate}
    \item \textbf{Initialization:} Sample \(N\) initial candidate steps \(s_1^1,\dots,s_1^N\).
    \item \textbf{Scoring:} Evaluate candidates based on verifier scores \(r(s_1^j \mid x)\).
    \item \textbf{Filtering:} Retain the top \(N/M\) candidates according to their scores.
    \item \textbf{Expansion:} For each retained candidate, generate \(M\) new candidate steps, resulting in \(N\) new candidate prefixes.
\end{enumerate}
Repeat steps 2–4 until a complete solution is reached. The final answer is selected via weighted BoN as described above.

\paragraph{Beam Search with Lookahead.}
Lookahead Search refines Beam Search by incorporating forward rollouts to enhance the verifier’s scoring accuracy at each step. Specifically, instead of directly using the verifier score at the current step, Lookahead Search performs a deterministic rollout up to \(k\) steps ahead (with temperature \(\tau=0\)) and scores the current step using:
\[
r(s_{k+1}\mid x, s_1,\dots,s_k),
\]
where \(s_2,\dots,s_{k+1}\) are generated deterministically from the model. Standard Beam Search is a special case with \(k=0\). This method closely mirrors a deterministic, exploitation-focused variant of Monte Carlo Tree Search (MCTS), removing stochastic exploration elements that are typically beneficial during training but less relevant at inference time.

\section{Additional Mathematical Details}

\subsection{Simplification of the Truncated Target Distribution}
\label{subsec:simplication-truncated}

In what follows we show that when the modulator is an indicator modulator 
\(f(z)=\mathbf 1\{z\ge\delta\}\), the target distribution simplifies to
the base-model distribution conditioned on the verifier event
\(r\ge\delta\); that is,
\[
p_{\text{target}}(\,\cdot \mid x,S_{<t}) = p_{\text{model}}(\,\cdot \mid x,S_{<t};\, r \ge \delta).
\]

By definition, one has
\begin{equation}
p_{\text{target}}(\,\cdot \mid x,S_{<t})
\;\propto\;
p_{\text{model}}(\,\cdot \mid x,S_{<t})\,\cdot
\mathbf{1}\bigl\{r(\,\cdot \mid x,S_{<t}) \ge \delta\bigr\}.
\end{equation}

Let
\begin{align}
Z
&:= \sum_{s\in\mathcal A^*} 
      p_{\text{model}}\bigl(s_t \mid x,S_{<t}\bigr)\, \cdot
      \mathbf{1}\!\bigl\{r(s \mid x,S_{<t}) \ge \delta\bigr\} \nonumber\\
&= \mathbb{P}\bigl(r(s_t\mid x,S_{<t}) \ge \delta\bigr).
\end{align}

where the probability is taken over $s_t\sim p_{\text{model}}$. Normalizing gives
\begin{equation}
p_{\text{target}}(s_t \mid x,S_{<t})
\;=\;
\frac{p_{\text{model}}(s_t \mid x,S_{<t})\,\cdot\mathbf 1\!\{r(s_t \mid x,S_{<t}) \ge \delta\}}
     {Z}.
\end{equation}

The right-hand side is precisely the model distribution conditioned on the event \(r \ge \delta\):
\begin{equation}
p_{\text{model}}\bigl(s_t \mid x,S_{<t};\, r \ge \delta\bigr)
\;=\;
\frac{p_{\text{model}}(s_t \mid x,S_{<t})\,\mathbf 1 \{r(s_t \mid x,S_{<t}) \ge \delta\}}
     {\mathbb{P}(r(s_t\mid x,S_{<t}) \ge \delta)}.
\end{equation}

Since the denominator equals \(Z\), we conclude
\begin{equation}
p_{\text{target}}(\,\cdot \mid x,S_{<t})
\;=\;
p_{\text{model}}(\,\cdot \mid x,S_{<t};\, r \ge \delta).
\end{equation}

\subsection{Connection Between LATTS and Local Difficulty}
\label{subsec:proof ar local difficulty}

In what follows, we show that, at each step, the expected number of candidate steps $s_t \sim p_{\text{model}}(\cdot \mid x, S_{<t})$ we need to sample before obtaining an accepted step $s_t \sim p_\text{target}$ scales with the local difficulty. In other words, steps with higher local difficulty require sampling more candidates.

Formally, let $n_t \in \mathbb{N}^*$ denote the number of candidates sampled before acceptance. Then, one has:
\begin{align}
    \mathbb{E}[n_t] 
    = \frac{1}{\mathbb{E}_{s_t \sim p_{\text{model}}(\cdot \mid x, S_{<t})}[r_f(s_t \mid x, S_{<t})]}
    = \frac{1}{1 - \Delta(x, S_{<t})}.
\end{align}

Indeed, each candidate step $s_t$ is sampled independently from the proposal distribution $p_{\text{model}}(\cdot \mid x, S_{<t})$ and is then accepted with probability $r_f(s_t \mid x, S_{<t})$. Thus, conditioning on having drawn a particular candidate $s_t$, the probability of accepting it is exactly $r_f(s_t \mid x, S_{<t})$.

To compute the overall probability of acceptance (without conditioning on a specific candidate), we marginalize over all possible candidates. By the law of total probability, this marginal (unconditional) acceptance probability is:
\begin{align}
\mathbb{P}(\text{accept}) 
= \mathbb{E}_{s_t \sim p_{\text{model}}(\cdot \mid x, S_{<t})}\bigl[\mathbb{P}(\text{accept}\mid s_t)\bigr]
= \mathbb{E}_{s_t \sim p_{\text{model}}(\cdot \mid x, S_{<t})}\bigl[r_f(s_t \mid x, S_{<t})\bigr].
\end{align}

This expectation corresponds precisely to the expected acceptance rate of candidates drawn from the proposal distribution, and by definition equals
\(1 - \Delta(x, S_{<t}) \).

Since each candidate is drawn independently with an identical acceptance probability, the sampling procedure at step $t$ corresponds exactly to repeated independent Bernoulli trials until the first success occurs. Thus, the number of candidates $n_t$ required follows a geometric distribution with success probability $1 - \Delta(x, S_{<t})$.

It immediately follows that the expected number of trials (candidate steps) before acceptance is given by the mean of this geometric distribution, providing the desired result:
\begin{align}
\mathbb{E}[n_t] = \frac{1}{1 - \Delta(x, S_{<t})}.
\end{align}

\subsection{Local Difficulty Notions Induced by Different Modulators}
\label{subsec:validity-ar}

As discussed in Section~\ref{sec:different-targets}, different choices of modulator $f$ induce different target distributions $p_\text{target} \propto p_\text{model} \cdot r_f$. Exploiting the connection between our acceptance--rejection scheme and the notion of local difficulty derived in Section~\ref{subsec:proof ar local difficulty}, different modulators $f$ lead to computational scaling at each step according to distinct notions of local difficulty, namely:

\begin{enumerate}[leftmargin=.8cm,itemsep=0cm,topsep=0cm,parsep=1pt]
    \item \textbf{LATTS Tilted:} Using $f(z) = z$ corresponds directly to the original verifier score, yielding the local difficulty measure:
    \[
    \Delta(x, S_{<t}) = 1 - \mathbb{E}_{s_t \sim p_{\text{model}}(\cdot \mid x, S_{<t})}[r(s_t \mid x, S_{<t})].
    \]
    This generalizes the completion-level difficulty concept introduced by \citet{lightman2023letsverifystepstep,snell2024scalingllmtesttimecompute} to individual reasoning steps, using the (raw) expected verifier score at each step.

    \item \textbf{LATTS Truncated:} Using $f(z) = \mathbf{1}\{z \geq \delta\}$ with threshold $\delta > 0$, we obtain:
    \[
    \Delta(x, S_{<t}) = 1 - \mathbb{P}\left(r(s_t \mid x, S_{<t}) \geq \delta\right) = \mathbb{P}\left(r(s_t \mid x, S_{<t}) < \delta\right),
    \]
    where the probability is computed under $p_{\text{model}}(\cdot \mid x, S_{<t})$. This quantifies the likelihood of generating an inadequate step (one scoring below $\delta$). To the best of our knowledge, this specific notion of local difficulty has not been previously explored, even at the completion level.
\end{enumerate}

\subsection{Validity of our Acceptance-Rejection Scheme}
\label{subsec:validity-ar}

Our AR approach satisfies the standard validity assumptions required by acceptance-rejection sampling. 

First, as our target distribution is defined by $p_\text{target} \propto p_\text{model} \cdot r_f$, it immediately follows that the support of $p_\text{target}$ is fully contained within the support of the proposal distribution $p_\text{model}$.

Second, acceptance-rejection sampling requires that the ratio between the target and the proposal distributions be bounded from above by a finite constant. In our setting, this ratio explicitly equals the modulated verifier score $r_f(s_t \mid x, S_{<t})$, which, by construction, lies within the interval $[0,1]$. Hence, the required boundedness condition is naturally satisfied with a bounding constant of $1$. 

Consequently, our AR sampling scheme is both theoretically justified and directly applicable.

\section{Chunk-wise Generation of Trials}
\label{sec:chunkwise-trials}

\begin{figure}[t]
    \centering
   \includegraphics[width=1\linewidth]
   {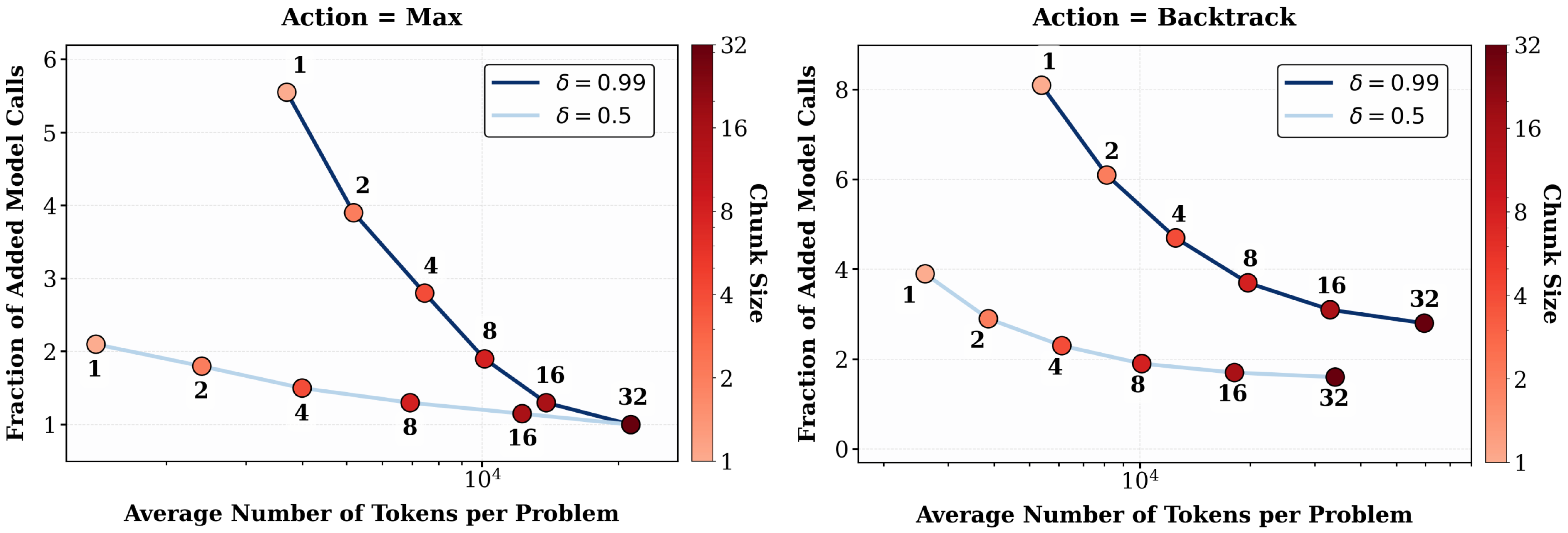}
   \caption{\textbf{Evaluating the effect of chunk size on latency to produce a solution.} We use Llama 3.2-Instruct-1B as the base model and Qwen2.5-7B PRM as the verifier, reporting results averaged over 10 completions per problem on the entire MATH500 dataset (totaling 5,000 completions). We consider two extremes of local difficulty using \( f(z)=\mathbf{1}\{z\geq\delta\} \) with thresholds \(\delta \in \{0.5,0.99\}\). Additionally, we analyze two fallback actions,\textsc{max} and \textsc{backtrack},triggered upon rejection of all candidate steps. As in other experiments, we fix the maximal number of trials to \(M=32\). We then report the FADM and and the average number of tokens generated per problem as a function of the chunk size $H$.}
\label{fig:chunk-wise-generation}
\end{figure}

Our method involves sampling and verifying candidate steps one at a time, which keeps token generation costs low, but may not be optimal for wall-clock latency, since it does not directly use the hardware's parallel processing capabilities.
However, when memory is available, we can easily modify our approach to use the following \emph{chunked} strategy that generates multiple candidates in parallel.

Let \(M\) be the maximum number of trials allowed for the current step and choose a chunk size
\(H\) such that \(1 \le H \le M\) and \(M \bmod H = 0\).
During each forward pass we proceed as follows:

\begin{itemize}[leftmargin=0.6cm,itemsep=0pt,topsep=2pt,parsep=0pt]
\item Draw a chunk of candidates
      \(s_t^{1},\dots,s_t^{H} \sim p_{\text{model}}(\cdot \mid x,S_{<t})\).
\item Sample i.i.d.\ thresholds
      \(u_1,\dots,u_H \sim \mathcal{U}([0,1])\).
\item Accept the candidate with the highest verifier score whenever
      \(\exists\,h\le H\) such that
      \(f\!\circ\! r(s_t^{h} \mid x,S_{<t}) \ge u_h\); set
      \(s_t \leftarrow
        \arg\max_{1 \le h \le H} r(s_t^{h} \mid x,S_{<t})\).
\item If no candidate is accepted, draw the next chunk.
      Continue until an acceptance occurs or \(M\) total trials have been exhausted.
\end{itemize}

If all \(M\) trials are rejected we invoke one of the fallback actions
described earlier,\textsc{stop}, \textsc{max}, \textsc{backtrack},
or \textsc{restart}.  This chunked scheme reduces latency by exploiting
parallelism while preserving the adaptive-computation property of
AR sampling.

\textbf{Remark.} When \( H = M \),that is, the chunk size equals the maximum number of trials,selecting the \textbf{max} fallback action upon rejection reduces our method to step-level BoN with \( N = M \).

We study how batching candidate generation in chunks of size \(H\) trades off
wall–clock latency against the total number of decoded tokens.
Latency is measured by the number of batched forward passes.
To quantify it, we report the \emph{Fraction of Added Model calls} (FADM)

\[
\text{FADM}
\;=\;
\frac{\texttt{num\_steps}+ \texttt{num\_resample}}{\texttt{num\_steps}},
\]

where \texttt{num\_steps} is the length of the reasoning chain and
\texttt{num\_resample} counts additional batched calls triggered by rejected
chunks; each chunk, whatever its size \(H>1\), is counted once.
Thus \(\text{FADM}=1\) means that every step is accepted at the
first attempt.

Larger \(H\) reduces latency,fewer forward passes are required,but increases the
total number of generated tokens.
Figure\;\ref{fig:chunk-wise-generation} illustrates the resulting trade-off.
Under the \textsc{max} fallback (continue with the best rejected candidate when a
chunk is fully rejected) FADM rapidly approaches \(1\) as \(H\) approaches the
per-step budget \(M=32\), recovering step-level Best-of-\(N\).
Even modest batching is effective:
for the acceptance threshold \(\delta=0.5\) we obtain \(\text{FADM}<2\) with \(H\ge 2\), while for the
stricter threshold \(\delta=0.99\) the same regime is reached with \(H\ge 8\). Under the \textsc{backtrack} fallback (discard the chunk and regenerate from the
previous step) FADM no longer converges to \(1\) as \(H\to M\), yet still yields
substantial latency gains. 
These results show that chunked sampling offers a flexible latency–compute
trade-off across verifier thresholds and fallback strategies.

\section{Algorithm}

The complete LATTS procedure, including our fallback strategy, is summarized in Algorithm~\ref{algo:latts}.

\newcommand{\INDSTATE}[1][1]{\STATE\hspace{#1\algorithmicindent}}
\begin{figure}
\centering
\begin{minipage}{0.9\linewidth}
    \begin{algorithm}[H]
        \caption{\textbf{LATTS}($p_\text{model}, r, f, M, \textsc{action}$)}
        \label{algo:latts}
        \begin{algorithmic}[1]
            \STATE \textbf{Require:} Base model $p_\text{model}$, verifier $r$, modulator $f$, maximal number of candidate trials per step $M$ (default $= 32$), fallback action $\textsc{action}$ (default $= \textsc{backtrack}$), maximal number of actions $A$ (default $= 8$).
            \STATE \textbf{Input:} Question $\EBGE{x}$.
            \STATE $\EGGE{S} \gets [\,\,]$ \hfill $\triangleright$ \textit{initialize reasoning chain}
            \STATE \textbf{while} generation not converged \textbf{do}:
            \INDSTATE $\ERED{\text{accepted}} \gets \text{False}$ \hfill $\triangleright$ \textit{reset acceptance flag}
            \INDSTATE $m \gets 0$,\,\,$a \gets 0$
            \INDSTATE \textbf{while} not $\ERED{\text{accepted}}$ and $m < M$ \textbf{do}:
            \INDSTATE[2] Sample $\EGGE{s_t^m} \sim p_\text{model}(\cdot \mid \EBGE{x}, \EGGE{S})$ \hfill $\triangleright$ \textit{sample candidate}
            \INDSTATE[2] Sample $u \sim \mathcal{U}([0,1])$ 
            \INDSTATE[2] \textbf{if} $u < f \circ r (\EGGE{s_t^m} \mid \EBGE{x}, \EGGE{S})$ \textbf{then}: \hfill $\triangleright$ \textit{acceptance-rejection}
            \INDSTATE[3] $\EGGE{S} \gets (\EGGE{S}, \EGGE{s_t^m})$ \hfill $\triangleright$ \textit{append candidate to reasoning chain}
            \INDSTATE[3] $\ERED{\text{accepted}} \gets \text{True}$
            \INDSTATE[2] $m \gets m + 1$
            \INDSTATE \textbf{if} not $\ERED{\text{accepted}} \text{ and } a  < A$ \textbf{then}: \hfill $\triangleright$ \textit{fallback action if all candidates are rejected}
            \INDSTATE[2] $\EGGE{S} \gets \textsc{action}(\EGGE{S}, \EGGE{s_t^1},\ldots,\EGGE{s_t^M})$ \hfill $\triangleright$ \textsc{stop}, \textsc{max}, \textsc{backtrack}, \textit{or} \textsc{restart} 
            \INDSTATE[2] $a \gets a + 1$
            \INDSTATE \textbf{else}: stop
            \STATE \textbf{Return} $\EGGE{S}$ 
        \end{algorithmic}
    \end{algorithm}
\end{minipage}
\end{figure}

\section{Additional Figures}
\label{sec:additional-figures}

\begin{figure}[h]
    \centering
   \includegraphics[width=0.9\linewidth]
   {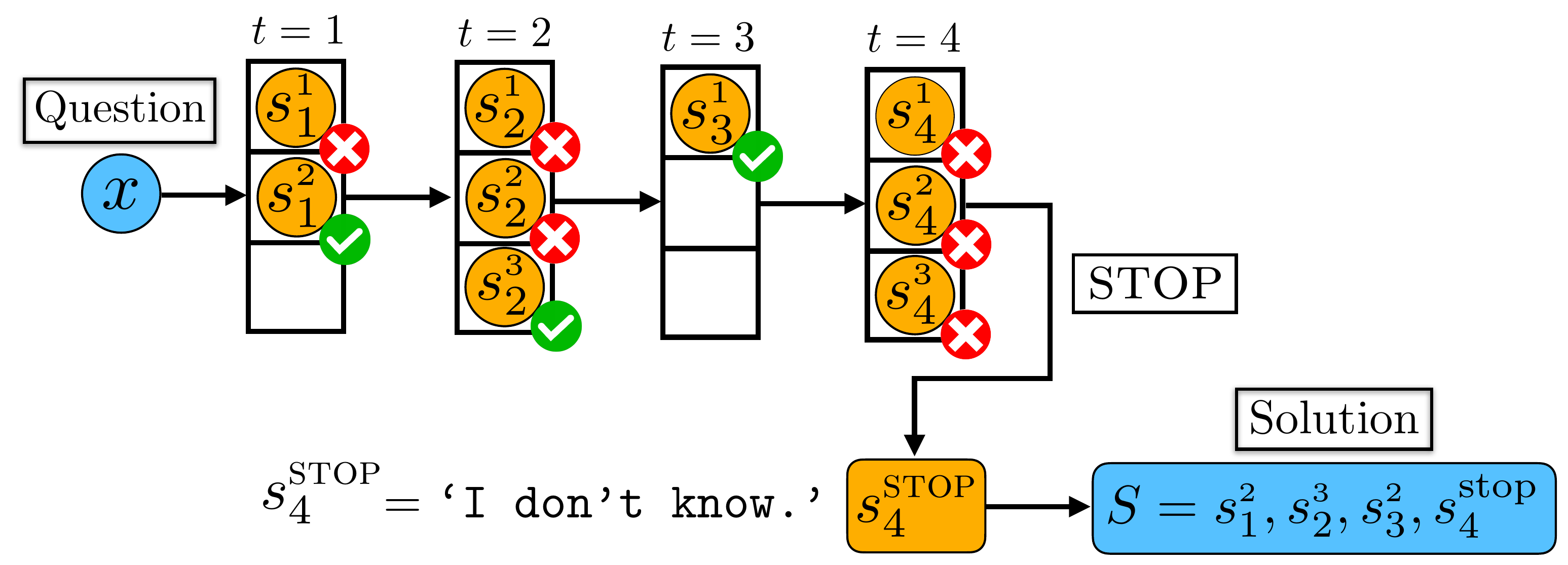}
   \caption{
   \textbf{LATTS with \textsc{stop} fallback.} After all four candidate steps are rejected at step 4, we halt the generation process, declare the solution invalid, and return the answer "I don't know."
}
\label{fig:latts_stop}
\end{figure}

\begin{figure}[h]
    \vspace{10mm}
    \centering
   \includegraphics[width=1\linewidth]
   {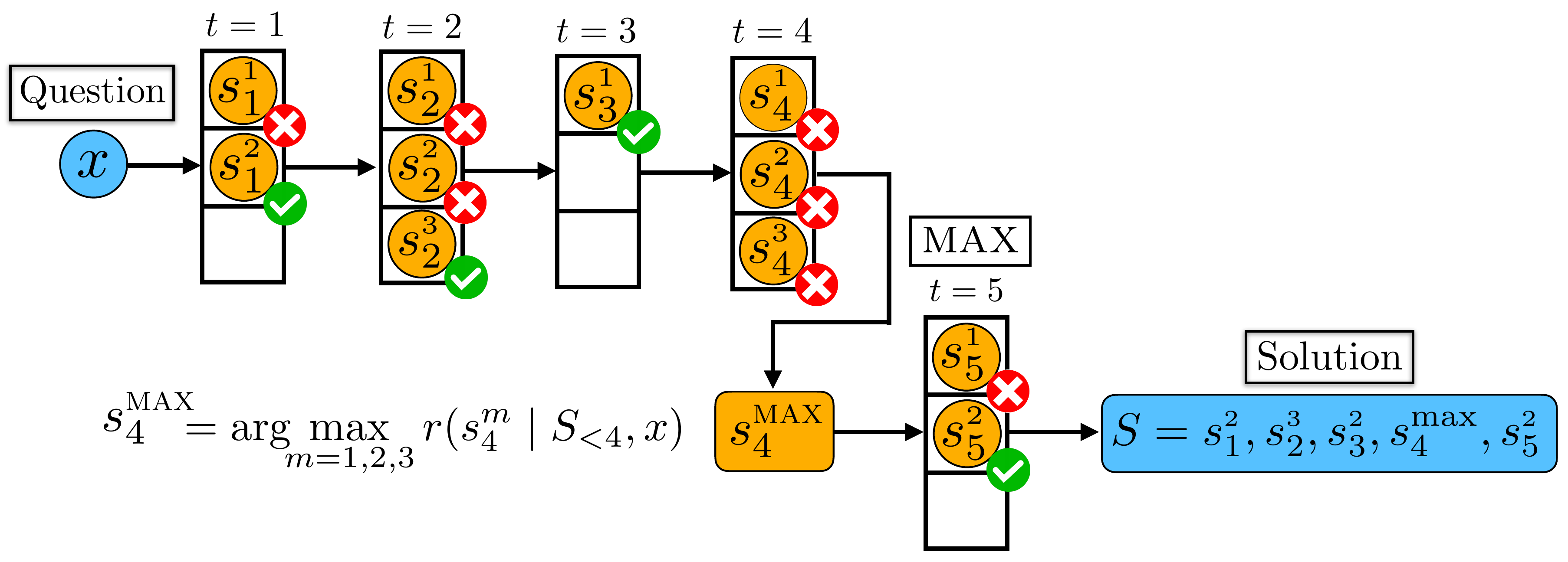}
   \caption{
   \textbf{LATTS with \textsc{max} fallback.} After all four candidate steps are rejected at step 4, we continue the generation process using the candidate that maximizes the verifier score (even though it was technically rejected).
}
\label{fig:latts_max}
\end{figure}

\begin{figure}[h]
    \vspace{-10mm}
    \centering
   \includegraphics[width=1\linewidth]
   {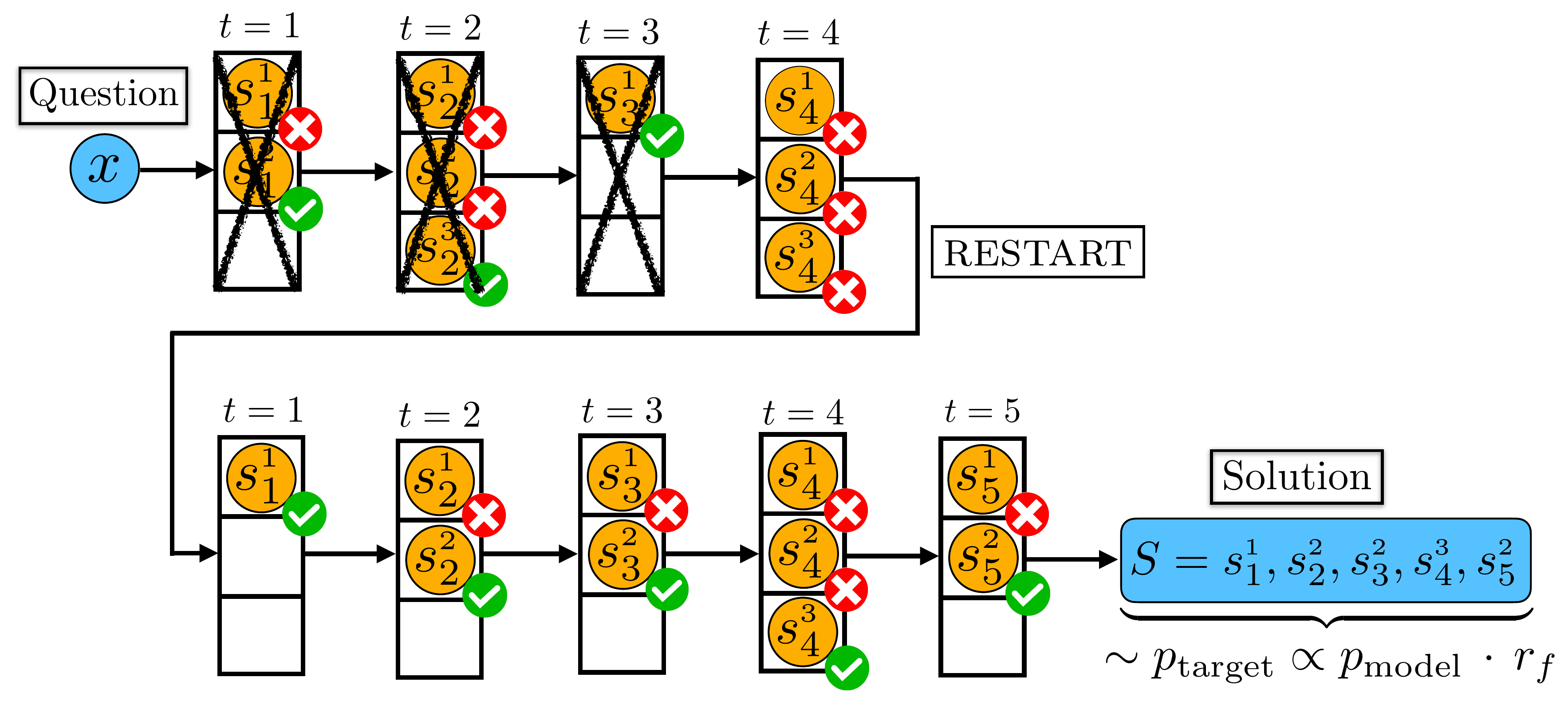}
   \caption{
   \textbf{LATTS with \textsc{restart} fallback}. After all four candidate steps are rejected at step 4, we restart the generation process.
}
\label{fig:latts_restart}
\end{figure}

In Figure~\ref{fig:latts_concept_figure},  we provide an illustrative example of LATTS with the \(\textsc{backtrack}\) fallback. In this section, we provide an illustrative example of LATTS with the other fallback strategies we consider, namely \(\textsc{stop}\) (Figure~\ref{fig:latts_stop}), \(\textsc{max}\) (Figure~\ref{fig:latts_max}) and \(\textsc{restart}\) (Figure~\ref{fig:latts_restart}). 

\section{LATTS Reasoning Traces}
\label{sec:reasoning-traces}

In this section, we illustrate the generation mechanism of LATTS through a concrete example of solving a problem from the MATH500 dataset. Specifically, we employ LATTS Truncated, which corresponds to LATTS using the modulator \( f = \mathbf{1}\{z \ge \delta\} \) with threshold \(\delta = 0.5\). Additionally, we use Qwen-2.5-PRM-7B~\citep{zhang2025lessonsdevelopingprocessreward} as the verifier. In this example, up to step 6, the first candidate is always accepted. However, at step 6, only the third candidate step is accepted, as the first two candidates contain calculation errors. Note that for this example, we never reach the maximal number of candidates, so the fallback action is never triggered.

\paragraph{Problem.} 

\begin{small}
\begin{verbatim}
If \( f(x) = \frac{3x-2}{x-2} \), what is the value of \( f(-2) + f(-1) + f(0) \)? 
Express your answer as a common fraction.
\end{verbatim}
\end{small}

\paragraph{Reasoning Trace.}
\begin{small}
\begin{Verbatim}[commandchars=\\\{\},numbers=left,firstnumber=\value{linenumber}]
Candidate:
## Step 1: Calculate f(-2)
To find f(-2), we plug x = -2 into f(x) = (3x-2)/(x-2).
f(-2) = (3(-2)-2)/(-2-2) = (-6-2)/(-4) = -8/(-4) = 2.
\textcolor{ForestGreen}{Verifier Score: 1.0}
\textcolor{ForestGreen}{Candidate accepted! Moving to the next step.}

Candidate:
## Step 2: Calculate f(-1)
To find f(-1), we plug x = -1 into f(x) = (3x-2)/(x-2).
f(-1) = (3(-1)-2)/(-1-2) = (-3-2)/(-3) = -5/(-3) = 5/3.
\textcolor{ForestGreen}{Verifier Score: 1.0}
\textcolor{ForestGreen}{Candidate accepted! Moving to the next step.}

Candidate:
## Step 3: Calculate f(0)
To find f(0), we plug x = 0 into f(x) = (3x-2)/(x-2).
f(0) = (3(0)-2)/(0-2) = (-2)/(-2) = 1.
\textcolor{ForestGreen}{Verifier Score: 1.0}
\textcolor{ForestGreen}{Candidate accepted! Moving to the next step.}

Candidate:
## Step 4: Calculate f(-2) + f(-1) + f(0)
Now, we add the results of f(-2), f(-1), and f(0):
f(-2) + f(-1) + f(0) = 2 + 5/3 + 1.
\textcolor{ForestGreen}{Verifier Score: 0.99609375}
\textcolor{ForestGreen}{Candidate accepted! Moving to the next step.}

Candidate:
## Step 5: Find a common denominator
To add the fractions, we need a common denominator, which is 3.
Convert 2 to a fraction: 2 = 6/3.
\textcolor{ForestGreen}{Verifier Score: 0.96484375}
\textcolor{ForestGreen}{Candidate accepted! Moving to the next step.}

Candidate:
## Step 6: Add the fractions
Now that we have the same denominator, we can add:
6/3 + 5/3 + 1 = 12/3.
\textcolor{red}{Verifier Score: 0.11328125}
\textcolor{red}{Candidate rejected. Not appending to the chain.}

Candidate:
## Step 6: Add the fractions
Now we can add:
(6/3) + (5/3) + 1 = 12/3.
Simplify the fraction: 12/3 = 4.
\textcolor{red}{Verifier Score: 0.06103515625}
\textcolor{red}{Candidate rejected. Not appending to the chain.}

Candidate:
## Step 6: Add the fractions
Correct addition:
(2 + 5/3 + 1) = (6/3 + 5/3 + 3/3) = 14/3.
\textcolor{ForestGreen}{Verifier Score: 0.9453125}
\textcolor{ForestGreen}{Candidate accepted! Moving to the next step.}

Candidate:
Final answer: \boxed{\frac{14}{3}}
\textcolor{ForestGreen}{Verifier Score: 1.0}
\textcolor{ForestGreen}{Candidate accepted! Moving to the next step.}

Final answer found. Stopping.
\end{Verbatim}
\end{small}
\setcounter{linenumber}{\numexpr\value{FancyVerbLine}+1\relax}

\section{Limitations}
\label{sec:limitations}

A limitation of our approach is that it is only applicable to domains 
where intermediate steps can be evaluated by a step-level verifier model. These constraints led us to evaluate the approach on mathematical problem-solving tasks, which naturally involve step-by-step reasoning, and have available specialized process reward models trained to verify step validity. However, the general method could potentially be applied more broadly on other domains where intermediate steps can be evaluated by a suitable verifier (e.g., scientific disciplines such as physics or chemistry where intermediate reasoning steps can be evaluated by a verifier, or potentially creative writing where sentences or paragraphs can be evaluated). Encouragingly, our experiments using a general-purpose LLM as a verifier instead of a specialized PRM (see Figure~\ref{fig:llm_as_critique}) suggest that a strong general-purpose model may replace a domain-specific step-by-step verifier, potentially enabling broader application to other domains.



\end{document}